\documentclass[conference]{IEEEtran}
\IEEEoverridecommandlockouts
\usepackage{cite}
\usepackage{amsmath,amssymb,amsfonts}
\usepackage{algorithmic}
\usepackage{graphicx}
\usepackage{textcomp}
\usepackage{xcolor}
\def\BibTeX{{\rm B\kern-.05em{\sc i\kern-.025em b}\kern-.08em
    T\kern-.1667em\lower.7ex\hbox{E}\kern-.125emX}}
\begin{document}

\title{A Reinforcement Learning based Path Planning Approach in 3D Environment\\
}

\author{
\IEEEauthorblockN{Geesara Kulathunga, }
\IEEEauthorblockA{Innopolis University}}

\maketitle

\begin{abstract}
Optimal motion planning involves obstacles avoidance where path planning is the key to success in optimal motion planning. Due to the computational demands, most of the path planning algorithms can not be employed for real-time based applications. Model-based reinforcement learning approaches for path planning have received certain success in the recent past. Yet, most of such approaches do not have deterministic output due to the randomness. We analyzed several types of reinforcement learning-based approaches for path planning. One of them is a deterministic tree-based approach and other two approaches are based on Q-learning and approximate policy gradient, respectively. We tested preceding approaches on two different simulators, each of which consists of a set of random obstacles that can be changed or moved dynamically. After analysing the result and computation time, we concluded that the deterministic tree search approach provides highly stable result. However, the computational time is considerably higher than the other two approaches. Finally, the comparative results are provided in terms of accuracy and computational time as evidence. 
\end{abstract}


%
\IEEEpeerreviewmaketitle

\section{Introduction}

Over the past decade, deep learning-enabled approaches have been obtained satisfactory results in various domains such as optimal motion planning, hyperspectral image classification, and path planning. In comparison to typical path planning algorithms, reinforcement learning (RL) based approaches were received significant attention in the recent past due to the success of the deep learning and computer vision. However, RL-based approaches without incorporating deep learning and computer vision were there for decades without proper success. 

Most RL-based approaches consist of V (value function) or Q (function), whose policies are defined in deterministic ways. However, the recent development of deep learning can help to incorporate sophisticated learning to formulate V or Q functions appropriately. RL-based path planning can be classified in two ways: model-base, model free. Model-based path panning approaches have several advantages over model-free approaches, for example, learning dynamics of quadrotor (model-free) is rather complicated because using a model-based approach does guarantee the dynamic feasibility since such approaches depend on the actual motion model. 

Typical graph-based or tree-based search algorithms, e.g., A*, Dijkstra and Min-Max, can be applied for planning, scheduling, discrete optimization, and games. In such search algorithms, in general, the algorithm describes by a set of states and actions that correspond to each state, which can be described explicitly in most situations. The objective of these algorithms is to find the best action at each state by expanding the tree or graph to minimize the cost or maximize the reward. However, tree or graph expansion becomes computationally expensive when state-space (states and actions combinations) complexity grows due to the enamours look ahead information. For example, state-space complexity of Reversi game is $10^{28}$, for Tic-tac-toe it is $10^3$, and for Go it is $10^{171}$ ~\cite{rocki2011large}. 

\textbf{Our main contribution} of the work is to explore the potentials of both RL-based path planning and deterministic-based path planning and how can incorporate the best of both to develop a fast and robust path planning approach, especially for quadrotors. For that, initially, we formulated the path planning problem for a deterministic system (actions are discrete and known) in which the optimal action in each state was selected by using the Optimal Planning of Deterministic Systems (OPD) algorithm~\cite{hren2008optimistic}.  

\begin{figure}[h]
    \centering
    \includegraphics[width=5cm]{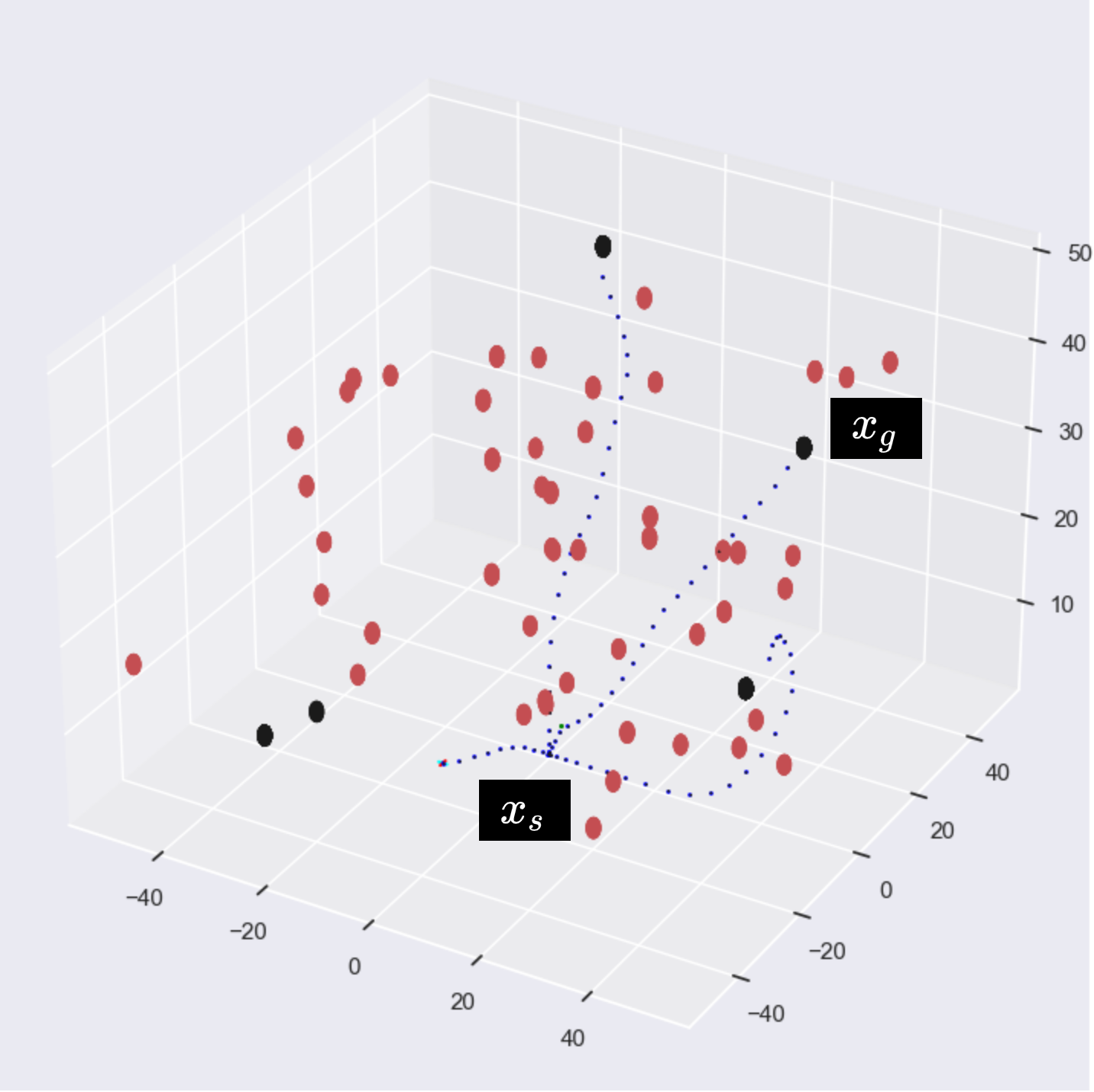}
    \caption{The improved tree based RL path planner. $x_s$ and $x_g$ denote start and goal positions, respectively. In blue colored trajectory depict the planned paths whose the start position kept the same position while changing the goal position }
    \label{fig:my_label}
\end{figure}

\section{Related work}

Motion planning can be carried out several different ways~\cite{paden2016survey}: search-based, sampling-based, and optimization-based. In search-based approaches, search space (or state space) is discretized into partitions, namely lattices and a path between the initial position and the end position are connected through a sequences of straight lines yields the shortest path between initial and end poses. Dijkstra's algorithm~\cite{barbehenn1998note}, A*~\cite{hart1968formal}, and D*~\cite{ferguson2005delayed} are a few widely used search-based algorithms. In most of the situations, real-time constraints are not met with search-based approaches due to high computational demands and memory footprints for state space representation. Sampling-based approaches do address state representation by incrementally drawing the state space.  RRT*~\cite{jaillet2008transition} and PRM~\cite{hsu2006probabilistic} are the mostly used ones. Optimization-based motion planning approaches can alleviate most of the problems preceding two approaches have. There are various techniques are proposed for this kind. Minimum-snap~\cite{mellinger2011minimum} one of the first proposed techniques for optimal motion planing for Quadrotors.  

In motion planning, agent (robot) is described by its state ($s_t \in S$) for a given time index where $S$ is the state space. Based on the reasoning approach, agent can take several different actions, e.g., $a_t \in A$ which lead agent to move to $s_{t+1} \in S$. In other word, let $P \in M(S)^{S\times A}$ be the system dynamics, $s_{t+1}$ is drawn based on a conditional probability, i.e., $P(s_{t+1}\mid s_t,a_t)$, where $M(S)$ denotes the a set of states within $S$. Moreover, control policy $\pi \in M(A)^S$, which agent's actions are drawn from, can be defined as $\pi(a_t\mid s_t)$. Hence, finding an optimal control policy $\pi^*$ is the main objective of agent at each time index t. There are various approaches can be applied for finding an $\pi^*$. Yet this is open research problem due to several reasons, including high computational demands, execution time, and the nature of the problem: linear or non-linear. In Reinforcement Learning (RL), finding an optimal control policy can be designed as a leaning-based decision making process, namely Marko Decision Process (MDP): ($S,A,P,R,\gamma$). Agent gets rewards $R(s_t,a_t)$, where $R\in[0,1]^{S\times A}$ when agent moving from $s_t$ to $s_{t+1}$. The ultimate objective is to maximize the rewards at each t, which eventually gives $\pi^*$. Thus, total reward is estimated by $R^{\pi} = \sum_{t=0}^{N}\gamma^tR(s_t, a_t)$, where N number of steps agent moves along a trajectory $\Pi = ((s_0,a_0), (s_1,a_1), (s_2,a_3),...,(s_{N},a_{N}))$. There could be exist multiple trajectories that lead to optimal solution. Hence, performance of a trajectory (policy $\pi$) is estimated through a value function, namely $V^{\pi}(s) = \mathbb{E}[R^{\pi} \mid s_0=s]$. Thus, objective is to find $V^{*}(s) = \max_{\pi}V^{\pi}(s), \; \forall s \in S$. 

Model-based and model-free are the main families in RL. Either of such families can be used for finding $\pi^*$. Yet, some of the work well over others due to problem characteristics. Model-based learning is preferred when the model dynamics are simple~(\cite{silver2016mastering,silver2017mastering,silver2018general}), and but $\pi$ estimation is complex. On the contrary, model-free learning is preferred for model with complex dynamics(\cite{degris2012model,glascher2010states}) yet $\pi$ estimation relatively simple. Dynamic Programming (DP) approaches (e.g., value iteration~\cite{bhowmik2010dynamic}, policy iteration~\cite{howard1960dynamic}) ensure the computational complexity around $O(\left | S \right |\left | A \right |log_{\gamma}(\epsilon (1-\gamma)))$ for discrete systems whose action-space is defined by $S\times A$. On the contrary, when S becomes continuous, Approximate Dynamic Programming (ADP) gives feasible solution over exact DP, where value function or policy is approximated within a provided hypothesis. Other than ADP, sampling-based optimization(or black-box optimizations) (e.g., Cross Entropy Method (CEM)\cite{amos2020differentiable}, Covariance Matrix Adaptation Evolution Strateg (CEM-ES)\cite{iruthayarajan2010covariance}) are used when S is large or continuous. Such approaches does not depends on the complete knowledge of MDP, but rather depends on the generator model, i.e., simulator, which can produce next state $s^` \sim P(s^` \mid s,a)$ and corresponding reword $R(s,a)$.

Monte Carlo Tree Search (MCTS)~\cite{chaslot2010monte} is preferred when the action space is discrete. MCTS has look-ahead three that starts from the current state and look for a optimal action to be taken as each step progressively expanding by sampling the trajectories using the generative model. The basic MCTS algorithm consists of four steps: selection, expansion, simulation and backpropagation. Selection function is applied to end of the current tree and one tree node is created in the expansion step. Take possible actions from such a expanded tree node and simulate till look-ahead node reach to its terminal state. In the backpropagation step, sum the total reward from the expansion node to terminal node is calculated by $v_i +  C \cdot \sqrt{\frac{ln N}{n_i}}$, where  $C$ is a exploration factor, $v_i$ mean value of node $n_i$, i.e., success ratio, $n_i$ is the number of times $n_i$ was visited and the total simulations count is given by $N$. The optimal or near-optimal action is identified using Probability Approximation Correct (PAC)~\cite{ekdahl2006bounds}. However, applicability for real-time applications is limited due to the high computational demands. In~\cite{hren2008optimistic} proposed one of the first polynomial regret bound algorithms which is quite faster, but applicability limited for deterministic dynamic and rewards. Later was extended~\cite{bubeck2010open} for supporting stochastic rewards and dynamics in a open-loop setting, i.e, sequence of actions. The preceding algorithm was further improved in~\cite{dfghfhf}. The modified version constraints on upper-confidence which helps to improve the performance considerably. For upper-bounds identification was determined with the help of Kullback-Leibler algorithm ~\cite{cappe2013kullback}. On the contrary, graph-based optimistic algorithms can capture across the states, which does not supports in the preceding algorithms (tree based algorithms). Such a algorithm was initially proposed by Silver et al.~\cite{silver2018general} where states values are derived from a shared Neural Network. Yet, authors of ~\cite{hostetler2014state} proposed state-space partitioning into smaller sets and aggregating similar states.  

The proximal policy optimization (PPO)~\cite{wang2020truly} is a supervised learning technique, where policy is defined by a deep neural network, e.g., multi layer perceptron, dueling network, convolutional network, ego attention network, and variational autoencoders~\cite{rivera2020anomaly}, depending on the situation. 

\section{Methodology}

\begin{figure}[h]
    \centering
    \includegraphics[width=6cm]{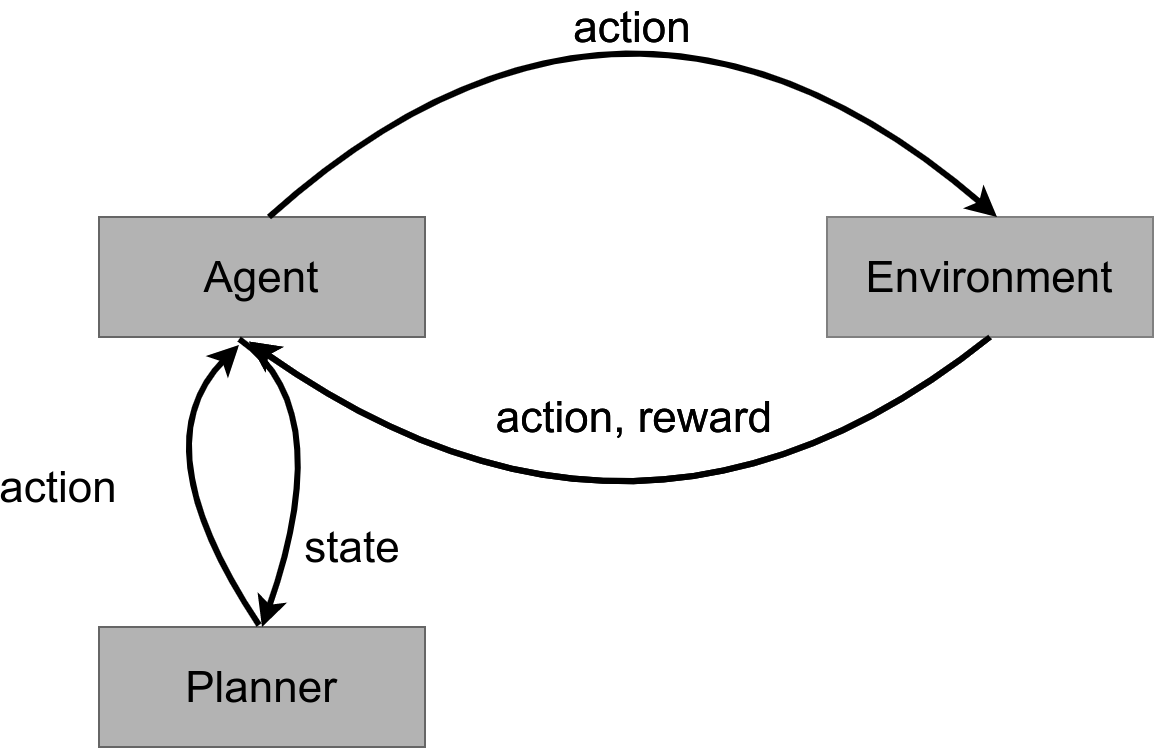}
    \caption{Other than true interaction between agent and environment, planning node simulates possible trajectories and pick the optimal action}
    \label{fig:action_planning}
\end{figure}

The deterministic control problem is defined as follows: $(X, A, f, r$, where action space and state space are denotes by A, X, respectively. Transition dynamics is given by $f: X \times A  \xleftarrow[]{} X $, and r denotes the reward function, which lies in the interval [0,1]. state space can be large yet action space should be deterministic, i.e., K. For any policy $\pi :X \xleftarrow{} A$, value function and Q-value function are defined by:
\begin{equation}
   \begin{aligned}
        V^{\pi}(x) = \sum_{t\geq 0} \gamma^tr(x_t, \pi(x_t)) \; 0 \leq \gamma < 1 \\
        Q^{\pi}(x,a) = r(x,a) + \gamma V^{\pi}(f(x,a))
   \end{aligned}
\end{equation} Thus, corresponding optimal V and Q functions can be determined as:
\begin{equation}
    \begin{aligned}
      V^{*}(x) = \max_{a \in A} r(x,a) + \gamma V^{*}(f(x,a)) \\
      Q^{*}(x,a) = r(x,a) + \gamma \max_{b\in A} Q^{*}(f(x,a), b)
    \end{aligned}
\end{equation} Let $A(n)$ be the action selection at each state, then regret yields selecting action from $A(n)$ over selecting optimal action, which can be formulated as:
\begin{equation}
    R_A(n) = \max_{a\in A} Q^{*}(x,a) - Q^{*}(x, A(n))
\end{equation}

As listed in the Related Work section, a variety of tree search algorithms for calculating optimal controlling polices have been suggested. However, due to high computational demands, most of them are not a feasible solution for real-time applications such as trajectory planning for quadrotors. For real-time planning, control policy must be recalculated at least at 5Hz to obtain a smooth flight experience. Hren and Munos~\cite{hren2008optimistic} proposed an algorithm, namely, OPD, that focuses on the the accuracy and computational demands at the same time for deterministic systems. The OPD uses the best possible resources (CPU time, memory) to solve the sequential decision making process. If execution is limited by time, the algorithm returns sub-optimal policy, otherwise, it returns higher accuracy policy. The OPD constructs a look-ahead tree at time instance t. In general, in order to explore a look-ahead tree, it is needed to get through all possible reachable states from the current state $x_t$. However, OPD proposed an optimal strategy for exploring the look-ahead tree from the current state $x_t$; thus, OPD works faster, which makes it more applicable for real-time applications. Three approaches of OPD application are Uniform look-ahead tree searching, Deep Q-Network, and Proximal Policy Optimization. They are detailed in further sections.

\subsection{Uniform look-ahead tree searching}
The foremost objective of  Uniform look-ahead tree searching is to retrieve an optimal action given a state $x$. In each state $x$, all possible reachable states are considered applying available actions. Such expanding new states are the children $c(i), i=0,...,K$ of the current state, when initial state is denoted as the root. The same process is applied recursively on each child to construct the next level of the tree. The order of the tree, d, is the number of the levels the tree $\beta$ grows. Hence, the cost of reaching any child $v_i$ of the tree is the supersum $v_i = max_{j\in C(i)}v_j$ and the optimal value of the root $v* = max_{i\in \beta} v_i = max \{v_1,...,v_K\}$. At any iteration, $\beta_n$ denotes the nodes that already expanded and $\beta_s = \beta - \beta_n$ denotes they have not expanded yet. Such unexpanded nodes are the possible nodes that can be expanded in the next iteration.  

\begin{figure}[h]
    \centering
    \includegraphics[width=6cm]{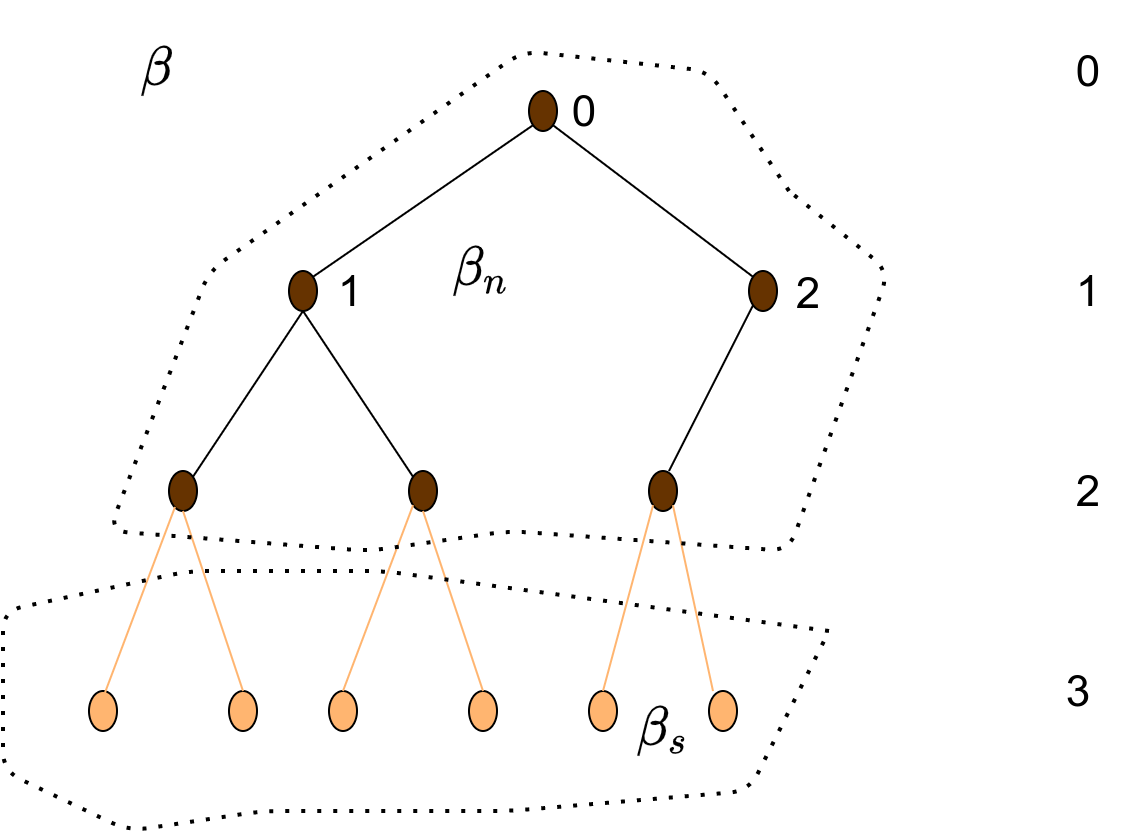}
    \caption{The regions $\beta_n$ and $\beta_s$ belong to expanded and to be expanded nodes. The number of children K is 2 at each node and tree depth d is 3}
    \label{fig:unifrom_tree_search}
\end{figure}

\subsection{Deep Q-Network}

Deep Q-Network (DQN) is the first deep reinforcement learning method proposed by DeepMind~\cite{mnih2013playing}. DQN can learn the control policies from the sensor data directly. Control policy is approximated from a deep neural network, e.g., convolutional neural network, whose input is constituted with the raw image and whose output is estimated using a value function, i.e., the reward function; the value function calculates the future action and uses the said action to estimate the control policy. Afterwards, the obtained control policy is learned by training a variant of Q-learning. The input for the network depends on several factors: robot environment, robot behaviour, and sensing capabilities of the robot. For example, in this research, we experimentally evaluated three different types of deep neural networks: multilayer perceptron, dueling network, and ego attention network for learning the control policy. For all the listed types, input for the network was obtained by the observations and distances to close-in obstacle positions after the normalization. Finally, the network output in DQN provides the probable action to be applied to the system.




\subsection{Proximal Policy Optimization}
The proximal policy optimization (PPO)~\cite{wang2020truly} is a supervised learning technique, which fill-in sampling data through interaction with the environment, that uses policy gradient methods for updating gradients, i.e, stochastic gradient ascent. Due the novel objective, i.e., $L^{clip}(\theta) = \hat{E}_t[min(r_t(\theta)\hat{A}_t), clip(r_t(\theta), 1-\epsilon, 1+\epsilon)\hat{A}_t]$, where $\theta$ is the policy parameter, $\hat{E}_t$ and $\hat{A}_t$ are the estimated expectation over timesteps and advantage, respectively, $\epsilon$ is a hyperparameter, and $r_t$ is the ratio of probability of new and old policies, that enable minibatch updates on multiple epochs. Moreover, PPO is able to update the trust region using trust region policy optimization (TRPO). Sampling complexity quite low due the preceding novel objective function.

\subsection{The problem formulation}

Despite the recent achievements in reinforcement learning, e.g., ~\cite{silver2018general}, most of the proposed solutions are not yet applied to industrial applications due to the undesirable behaviour. However, model-based learning approaches can alleviate such undesirable behaviours often better than their model-free counterparts. Elouard~\cite{leurent2020robust} introduced an arbitrary reward function R which can be maximized to shortages that come in critical setting where mistakes that are made from the model to be minimized. More precisely, consider the following linear system
\begin{equation}
\begin{aligned}
  \dot{x}(t) = A(\theta) x(t) + Bu(t) +D\epsilon(t), \; \epsilon(t) \geq 0, \\
  y(t) = \dot{x}(t) + Cz(t)
\end{aligned}
\end{equation} where $\theta$ in the state matrix $A(\theta) \in \mathbb{R}^{n_x\times n_x} \subset \Psi \in \mathbb{R}^d$, control matrix $B \in \mathbb{R}^{n_x\times n_u}$, disturbance matrix $D \in \mathbb{R}^{n_x\times n_u}$. $x \in \mathbb{R}^n_x$, $u \in \mathbb{R}^{n_u}$ and $\epsilon \in \mathbb{R}^{n_r}$ are state, control and disturbances, respectively. The system observation is given by $y(t)$, where $z(t) \in \mathbb{R}^{n_n}$ and $C \in \mathbb{R}^{n_x \times n_n}$. B, C, D are known matrices with respect to the considered system. Afterwards, they defined a robust control objective aims to maximize the worse-case next state output with respect to a confident region $C_{n, \delta}$
\begin{equation}
    \max_{u \in \mathbb{R}^{\mathbb{N}}} V^{r}(\mathbf{u})
\end{equation}, where $V^{r}(\mathbf{u}) = \underset{\theta \in C_{N, \delta}}{\inf }[\Sigma_{n=N+1}^{\infty} \gamma^{n}R(x_n(\mathbf{u}, \omega))], \; \gamma \in (0,1) $.  Since system is linear A is known, $A(\theta)$ is estimated as follows:

\begin{equation}
    A(\theta) = A + \Sigma_{i=1}^d \theta_i \pi_i
\end{equation} More information regarding the $\theta$ estimation~\cite{leurent2020robust}. 
Simulator we developed for avoiding the obstacles in 3D space from start position to desired goal position.  The system is described by  $\mathbf{x} =[ p_x \; p_y  \; p_z  \; v_x  \; v_y  \; v_z]^T$, where $p_\mu, v_\mu, \mu = x,y,z$ are position and velocity along the $\mu$ direction.  Action space $u=(u_x, u_y, u_z) \in [-1,1]^3$ is divided into 8 directions: forward, backward, left, right, top, and down, as follows:
\begin{equation*}
\begin{aligned}
  A = \{(0,0,0), (1,0,0),(-1,0,0),(0,1,0),(0,-1,0),\\(0,0,1),(0,0,-1)]\}
\end{aligned}
\end{equation*} The reward encodes cost towards to desired goal state $x_g$, while avoiding collisions with obstacles:
\begin{equation}
\begin{aligned}
     R(x) = \delta(x)/(1 + \|x - x_g\|_2), \\
     \delta(x) = \left\{\begin{matrix}
0 & \delta(x)< obs_d \\ 
 \delta(x)  & otherwise
\end{matrix}\right.,
\end{aligned}
\end{equation} where $obs_d$ denotes the minimum acceptance distance, i.e., 1m. The system dynamics consist  with friction parameters $(\theta_x, \theta_y, \theta_z)$:
$$
\begin{bmatrix}
\dot{p_x}\\
\dot{p_y}\\
\dot{p_z}\\
\dot{v_x}\\
\dot{v_y}\\
\dot{v_z}\\
\end{bmatrix} = 
\begin{bmatrix}
0 & 0 & 0 & 1 & 0 & 0\\
0 & 0 & 0 & 0 & 1 & 0\\
0 & 0 & 0 & 0 & 0 & 1\\
0 & 0 & 0& -\theta_x & 0 &0 \\
0 & 0 & 0  & 0 & -\theta_y &0 \\
0 & 0 & 0  & 0 & 0 & -\theta_z \\
\end{bmatrix}
\begin{bmatrix}
{p_x}\\
{p_y}\\
{p_z}\\
{v_x}\\
{v_y}\\
{v_z}\\
\end{bmatrix}
+
\beta \begin{bmatrix}
0\\
0\\
0\\
{u_x}\\
{u_y}\\
{u_z}\\
\end{bmatrix},
$$ where $0 < \beta \leq 1 $ is a scaling parameter that control the direction of the action. 

\begin{figure}[]
    \centering
    \includegraphics[width=5cm]{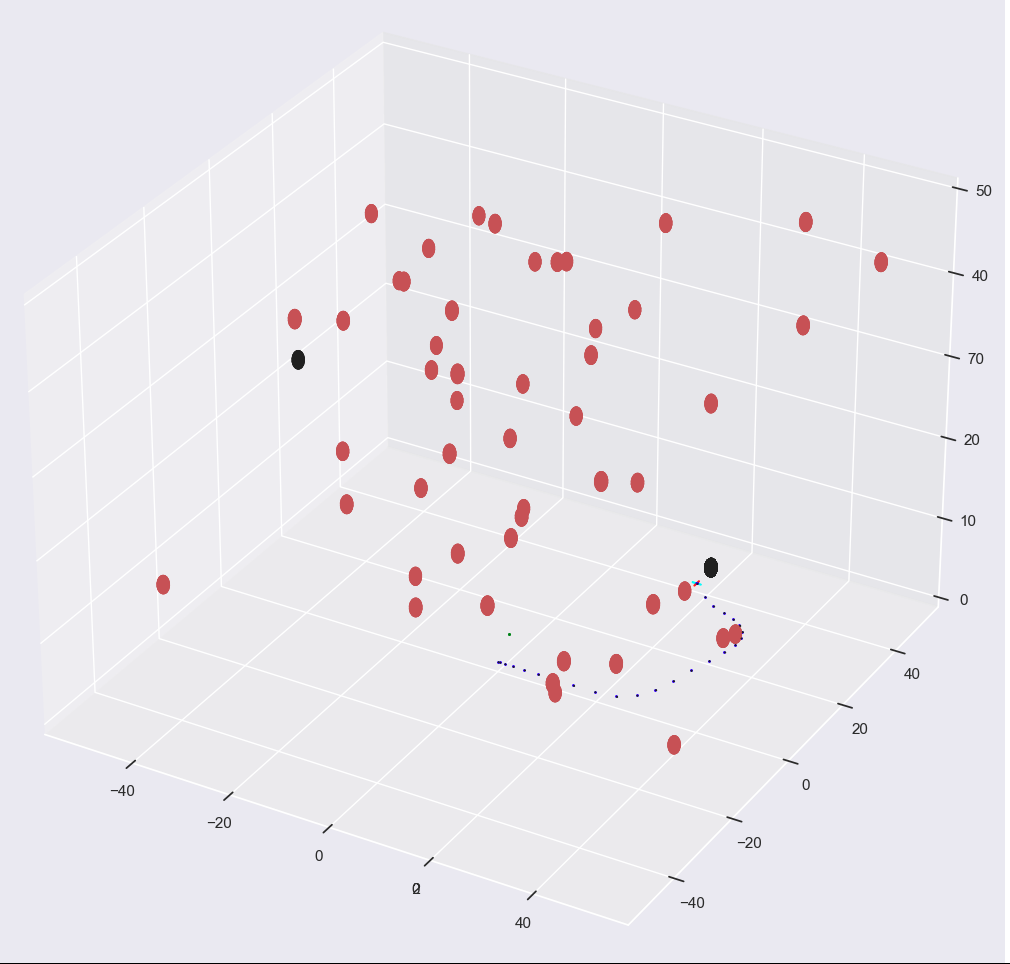}
    \caption{Back and red spheres denote the goal positions and obstacles positions, respectively. The quadrotor starts at the origin (0,0,0) moving towards to goal position while avoiding the obstacles}
    \label{fig:my_label}
\end{figure}

\section{Experimental Results}
We have modified the approach that proposed in~\cite{leurent2020robust}, supporting for 3D environment. In the improved version we incorporated OctoMap based map representation where obstacle-free distance can be calculated from a desired position. Such a distance estimation was helped to faster convergence of look-ahead tree search, while keeping the almost the same accuracy. We have ran theirs with the proposed over 100 episodes and calculated mean execution time over as given in Table~\ref{tab:_line}. Furthermore, we have tried with Proximal Policy Optimization with three different control policy generation, i.e.,  Convolutional Neural Network, Variational Autoencoder, and Dueling Neural Network. However, we have not achieved high accurate trajectory when we used Proximal Policy Optimization for predicting of next optimal action.  Success fraction is ratio of the number of successful trajectories to total number of tries, i.e., 100 times. Mean execution time is calculated total time is defined as ratio of time utilized for the total trajectory calculation to number of steps. With reference to Table.~\ref{tab:_line}, it is clear that the proposed approach has the optimal in comparison to the other approach that have compared.  

\section{Conclusion and Discussion}
Accuracy, computation time, and memory consumption make path planning a challenging task for real time application. Current advancement in the field do not allow mobile robots, e.g., drones and autonomous ground vehicles, to handle dynamic obstacles robustly. Two main classes of approaches that have been used to address the said constrains are RL-based and deterministic-based approaches. In this work, we have compared three RL-based approaches and one deterministic approach for the same path planning task, i.e., finding an obstacle-free path between start and goal position. Then, we proposed a hybrid approach that combines Monticalo tree search and RL-based approach to solve the same task. In terms of reaching the goal position, the accuracy of the proposed hybrid approach is considerably higher than the other three approaches: Proximal policy optimization, deep Q-network, and uniform tree search; this ensures the safety and dynamic feasibility.  
However, computational demands of the proposed hybrid approach are sightly higher compared to the aforementioned three approaches, which constitutes a limitation of the proposed hybrid approach because such demands yield ineffability to apply in real-world motion planning applications. Thus, the next stage of this research project focuses on incorporating Proximal Policy Optimization-based planning to further reduction of computational time while keeping the high accuracy. Further research could also focus on memory footage of the algorithm.

\begin{table}[]
\caption{Comparison the proposed approach with a few other variants}

\centering
\begin{tabular}{|l|l|}
\hline
\textbf{Approach} & \textbf{Control Policy  Estimation} \\ \hline
\cite{leurent2020robust} & Deterministic Tree Search \\ \hline
Improved version of \cite{leurent2020robust}& \begin{tabular}[c]{@{}l@{}}Uniform look-ahead \\ Tree Search\end{tabular} \\ \hline
Proximal Policy Optimization & \begin{tabular}[c]{@{}l@{}} Convolutional Neural\\  Network\end{tabular} \\ \hline
Proximal Policy Optimization & Variational Autoencoder  \\ \hline
Proximal Policy Optimization & Dueling  Neural Network \\ \hline
\textbf{Mean Execution Time (s)} & \textbf{Success Fraction} \\ \hline
\multicolumn{1}{|c|}{2.234} & \multicolumn{1}{c|}{0.967} \\ \hline
\multicolumn{1}{|c|}{\textbf{1.230}} & \multicolumn{1}{c|}{ \textbf{0.912}} \\ \hline
\multicolumn{1}{|c|}{0.345} & \multicolumn{1}{c|}{0.235} \\ \hline
\multicolumn{1}{|c|}{0.334} & \multicolumn{1}{c|}{0.342} \\ \hline
\multicolumn{1}{|c|}{0.315} & \multicolumn{1}{c|}{0.453} \\ \hline
\end{tabular}
\label{tab:_line}
\end{table}

\bibliographystyle{IEEEtran}

\bibliography{references}

\end{document}